# Impact of Exponent Parameter Value for the Partition Matrix on the Performance of Fuzzy C Means Algorithm

Dibya Jyoti Bora[1]     ,Dr. Anil Kumar Gupta[2]
[1] *Department Of Computer Science and Applications, Barkatullah University, Bhopal, India*
[2] *Department Of Computer Science and Applications, Barkatullah University, Bhopal, India*

**Abstract-**Soft Clustering plays a very important rule on clustering real world data where a data item contributes to more than one cluster. Fuzzy logic based algorithms are always suitable for performing soft clustering tasks. Fuzzy C Means (FCM) algorithm is a very popular fuzzy logic based algorithm. In case of fuzzy logic based algorithm, the parameter like exponent for the partition matrix that we have to fix for the clustering task plays a very important rule on the performance of the algorithm. In this paper, an experimental analysis is done on FCM algorithm to observe the impact of this parameter on the performance of the algorithm.
***Keywords*-Clustering, FCM, Matlab, Soft clustering**

## I. INTRODUCTION

Data clustering is an unsupervised study where we try to divide our data into separate groups known as "clusters". But on doing this, we have to maintain two important properties that are: 1. High Cohesive Property and 2. Low coupling property. According to the first property, the data items inside a cluster should exhibit high similar properties, and the second property says that data items inside a cluster must be different in nature from the data items in a different cluster. The clustering field is again divided into two categories: 1. Exclusive clustering (Hard Clustering) & 2. Overlapping clustering (Soft clustering).

In case of hard clustering, a data item must exclusively belong to only one cluster. Among the popular ones, K-Means is a frequently chosen algorithm for hard clustering task. While, in case of soft clustering task, a membership value is assigned to every object based on which an object may simultaneously belong to more than one cluster. FCM is a popular soft clustering technique. In this paper, first an introduction of soft clustering and FCM algorithm is given, and then we go for an experimental analysis of FCM algorithm by observing the impact of different values of exponent for the partition matrix on the performance of the algorithm.

## II. SOFT CLUSTERING

Soft clustering is also referred as overlapping clustering or fuzzy logic based clustering [1].In soft clustering, a data item may not exclusively belong to particular one cluster. Depending on the membership value, it may belong to more than one cluster. Some times, hard clustering is not suitable for our clustering task. As for example: In *Each Movie* dataset used to test recommender systems [2], many movies belong to more than one genre, such as "Aliens", which is listed in the action, horror and science fiction genres. Similarly, in case of document categorizations, a single document may belong to more than one category [3]. In such cases, soft clustering is chosen for clustering task. In soft clustering, a data item is associated with a set of membership levels which indicates the strength of the association between that data element and a particular cluster [4]. Fuzzy logic is used in the involved mathematical calculations.

## III. FUZZY C-MEANS ALGORITHM

*Fuzzy C-means* (FCM) is a soft data clustering technique wherein each data point belongs to a cluster to some degree that is specified by a membership grade. This technique was originally introduced by Jim Bezdek in 1981 as an improvement on earlier clustering methods [4] [5]. It provides a method of how to group data points that populate some multidimensional space into a specific number of different clusters. The main advantage of fuzzy c – means clustering is that it allows gradual memberships of data points to clusters measured as degrees in [0,1]. This gives the flexibility to express that data points can belong to more than one cluster.

It is based on minimization of the following objective function:

$$J_m = \sum_{i=1}^{N} \sum_{j=1}^{C} u_{ij}^m \|x_i - c_j\|^2$$

$$1 \leq m < \infty$$

where *m* is the value of the exponent which is any real number greater than 1, $u_{ij}$ is the degree of membership of $x_i$ in the cluster *j*, $x_i$ is the $i^{th}$ of d-dimensional measured data, $c_j$ is the d-dimension center of the cluster, and ‖*‖ is any norm expressing the similarity between any measured data and the center.

Fuzzy partitioning is carried out through an iterative optimization of the objective function shown above, with the update of membership value $u_{ij}$ and the cluster centers $c_j$ by:

$$u_{ij} = \frac{1}{\sum_{k=1}^{C}\left(\frac{\|x_i - c_j\|}{\|x_i - c_k\|}\right)^{\frac{2}{m-1}}}$$

$$c_j = \frac{\sum_{i=1}^{N} u_{ij}^m \cdot x_i}{\sum_{i=1}^{N} u_{ij}^m}$$

This iteration will stop when $\max_{ij}\left\{\left|u_{ij}^{(k+1)} - u_{ij}^{(k)}\right|\right\} < \varepsilon$, where $\varepsilon$ is a termination criterion between 0 and 1, whereas $k$ are the iteration steps. This procedure converges to a local minimum or a saddle point of $J_m$.

The formal algorithm is:

1. *Initialize $U=[u_{ij}]$ matrix, $U^{(0)}$*
2. *At k-step: calculate the centers vectors $C^{(k)}=[c_j]$ with $U^{(k)}$*

$$c_j = \frac{\sum_{i=1}^{N} u_{ij}^m \cdot x_i}{\sum_{i=1}^{N} u_{ij}^m}$$

3. *Update $U^{(k)}$, $U^{(k+1)}$*

$$u_{ij} = \frac{1}{\sum_{k=1}^{C}\left(\frac{\|x_i - c_j\|}{\|x_i - c_k\|}\right)^{\frac{2}{m-1}}}$$

4. *If $\| U^{(k+1)} - U^{(k)} \| < \varepsilon$ then STOP; otherwise return to step 2.*

The time complexity of FCM algorithm is $O(ndc^2i)$ [9][11], where n is the number of data points, c is the number of clusters, d is the dimension of the data and i is the number of iterations. In this paper, our research is focusing mainly on the value of exponent for the partition matrix. Because, the updating criteria of FCM is determined by this parameter value. This exponent value determines the degree of fuzziness. Generally, it is assigned a value greater than 1. When the exponent value is tending to infinity, the fuzzy degree is increasing. In the next section, we will experimentally analyze the FCM algorithm with different values exponent for the partition matrix in Matlab.

## IV. EXPERIMENTAL RESULTS

We have chosen Matlab for our experiments. In Matlab, Fuzzy C Means clustering can be performed with the function "fcm". This function can be described as follows [6][10]:

*[center, U, obj_fcm] = fcm(data,cluster_ n)*

The arguments of this function are:

1) *data* - lots of data to be clustering, each line describes a point in a multidimensional feature space;
2) *cluster_n* - number of clusters (more than one).

The function returns the following parameters:

1) *center* - the matrix of cluster centers, where each row contains the coordinates of the center of an individual cluster;
2) *U* - resulting matrix;
3) *obj_fcn* - the objective function value at each iteration

For, our experiment, we have chosen "iris" dataset [7][8][9]. The Iris flower data set or Fisher's Iris data set (some times also known as Anderson's Iris data) is a multivariate data set introduced by Sir Ronald Fisher (1936) as an example of discriminant analysis. It consists of 50 samples from each of three species of Iris (Iris setosa, Iris virginica and Iris versicolor). Four features were measured from each sample: the length and the width of the sepals and petals, in centimeters [7]. Table (1) gives details of the iris dataset [7]:

| **Data Set Characteristics:** | Multivariate | **Number of Instances:** | 150 | **Area:** | Life |
|---|---|---|---|---|---|
| **Attribute Characteristics:** | Real | **Number of Attributes:** | 4 | **Date Donated** | 1988-07-01 |
| **Associated Tasks:** | Classification | **Missing Values?** | No | **Number of Web Hits:** | 548538 |

Table (1): Iris dataset

We have initialized the required number of clusters as 4, minimum improvement factor as 1e-6 and maximum iterations count as 100. Now, starting from the exponent value = 1.5, we have run the FCM algorithm for the same clustering task of the iris dataset with different values of exponent as 1.6, 1.7, 1.8, 1.9 and 2.0. We have observed the following results (Diagrams are shown with respect to each observation):

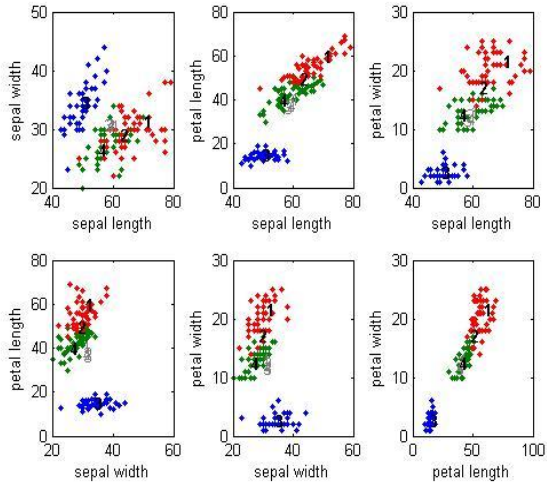

Figure [1]: Clustering with respect to expo = 1.5

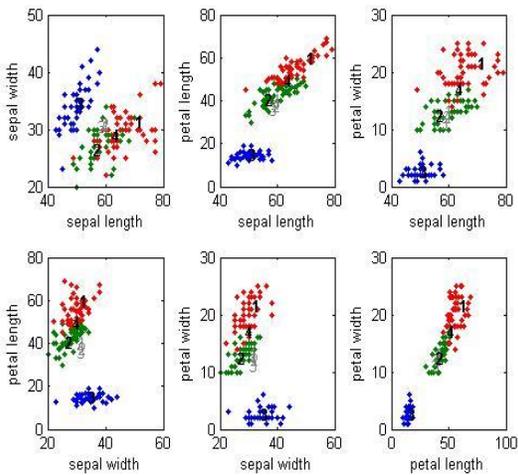

Figure [2]: Clustering with respect to expo = 1.6

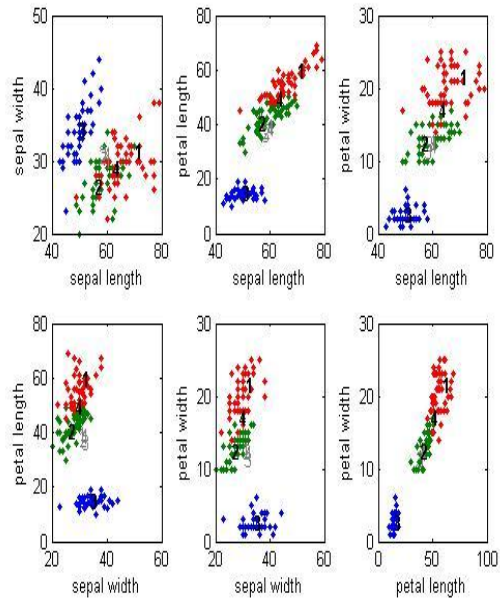

Figure [3]: Clustering with respect to expo = 1.7

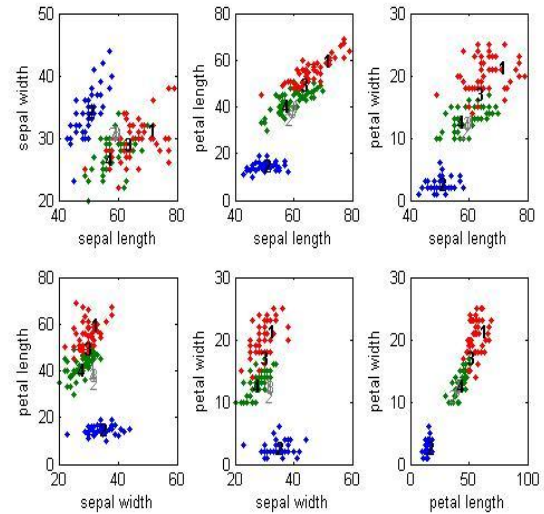

Figure [4]: Clustering with respect to expo = 1.8

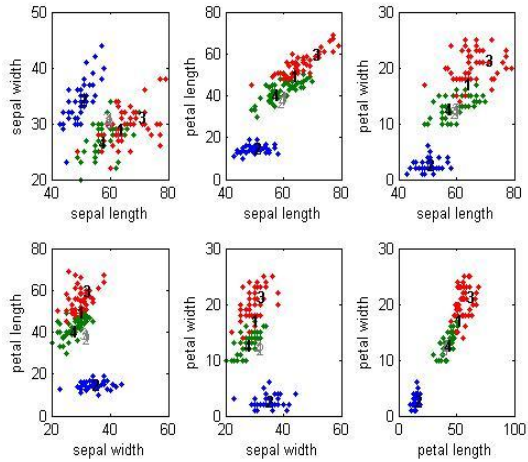

Figure [5]: Clustering with respect to expo = 1.9

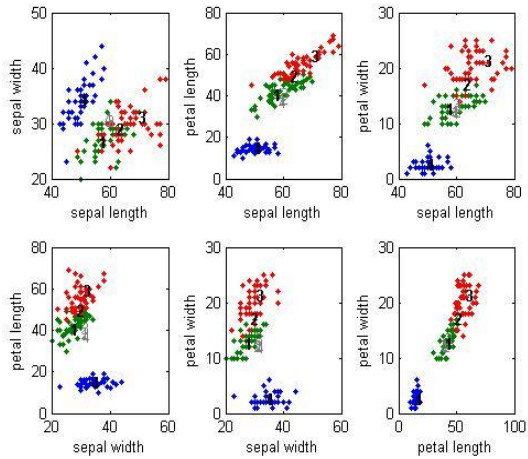

Figure [6]: Clustering with respect to expo = 2.0

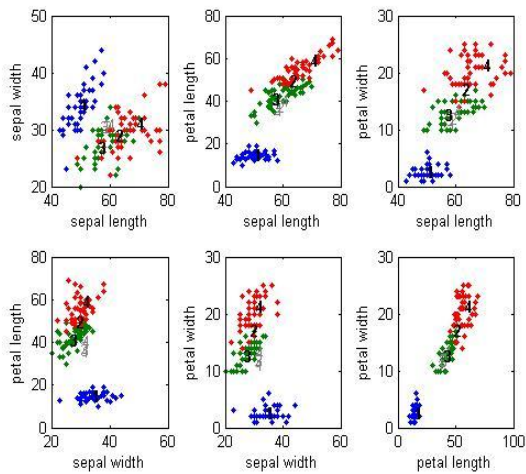

Figure [7]: Clustering with respect to expo = 2.1

The overall results are shown below in the Table (2):

| Exponent Value | Maximum Iteration Count | Value of The Objective Function | Time Taken(in seconds) |
|---|---|---|---|
| 1.5 | 49 | 5384.478684 | 1.7188 |
| 1.6 | 47 | 5203.513296 | 2.5938 |
| 1.7 | 45 | 4984.149622 | 1.0938 |
| 1.8 | 46 | 4733.021835 | 1.2031 |
| 1.9 | 45 | 4458.383501 | 1.1875 |
| 2.0 | 39 | 4168.707060 | 1.0469 |
| 2.1 | 36 | 3871.837094 | 1.0000 |

Table (2): Experimental results with respect different exponent values

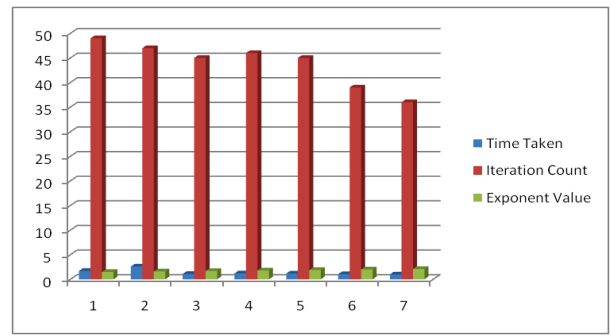

Figure [8]: Time taken and Maximum Iteration Count for Different Exponent Values

So, it is clear from the above experimental result that for the exponent value 2.1, FCM algorithm is giving better performance in terms of minimum time, minimum no of maximum iteration count and minimum objective function value.

V. CONCLUSION

In this experimental study, we have found that the exponent value 2.1 for the partition matrix in FCM algorithm is well suitable for clustering iris dataset. We have observed that the value of the exponent for the partition matrix has full impact on the performance of the FCM algorithm. Actually, it is the parameter who determines the fuzziness of the algorithm. Choosing a good exponent value will obviously result a better clustering performance. But it is really difficult to determine a perfect value for the exponent parameter. So, our future research work will purely focus on this fuzzy parameter means how to select a better exponent value for the partition matrix and what criteria we have to consider for this purpose. In fact, property of dataset is also a major

2319 – 1953

criterion that we always have to pay notice on before going for clustering task. So, in future, we'll analyze our experiment for different datasets with varying properties.